\documentclass[letterpaper, 10 pt, conference]{ieeeconf}  

\IEEEoverridecommandlockouts                              

\usepackage{graphics}
\usepackage{algorithm}
\usepackage{algorithmic}
\usepackage{booktabs}
\graphicspath{{figures/}}

\usepackage{amsmath,amsfonts,bm}

\def\eqref#1{equation~\ref{#1}}

\def\1{\bm{1}}

\DeclareMathAlphabet{\mathsfit}{\encodingdefault}{\sfdefault}{m}{sl}
\SetMathAlphabet{\mathsfit}{bold}{\encodingdefault}{\sfdefault}{bx}{n}

\usepackage{xcolor}
\usepackage{multirow}
\usepackage{amssymb}
\usepackage{color}
\usepackage{adjustbox}
\usepackage{cite}
\usepackage{url}

\newcommand{\pa}{\textcolor{black}}

\def\etal{\emph{et al}.}

\title{\LARGE \bf
DMODE: Differential Monocular Object Distance Estimation Module without Class Specific Information
}

\author{Pedram Agand$^{1}$, Michael Chang$^{1}$, and Mo Chen${^1}$
\thanks{$^{1}$Simon Fraser University, Burnaby, Canada
        {\tt\small \{pagand, michael\_chang\_7 mochen\}@sfu.ca}}%
}

\begin{document}

\maketitle
\thispagestyle{empty}
\pagestyle{empty}

\begin{abstract}
Utilizing a single camera for measuring object distances is a cost-effective alternative to stereo-vision and LiDAR. Although monocular distance estimation has been explored in the literature, most existing techniques rely on object class knowledge to achieve high performance. 
Without this contextual data, monocular distance estimation becomes more challenging, lacking reference points and object-specific cues. However, these cues can be misleading for objects with wide-range variation or adversarial situations, which is a challenging aspect of object-agnostic distance estimation. In this paper, we propose DMODE, a class-agnostic method for monocular distance estimation that does not require object class knowledge. DMODE estimates an object's distance by fusing its fluctuation in size over time with the camera's motion, making it adaptable to various object detectors and unknown objects, thus addressing these challenges. We evaluate our model on the KITTI MOTS dataset using ground-truth bounding box annotations and outputs from TrackRCNN and EagerMOT. The object's location is determined using the  change in bounding box sizes and camera position without measuring the object's detection source or class attributes. Our approach demonstrates superior performance in multi-class object distance detection scenarios compared to conventional methods. 
\end{abstract}

\section{INTRODUCTION}
\label{sec:intro}

For AI-enabled object detection, applications in simultaneous localization and mapping (SLAM), virtual reality, surveillance video perception and autonomous vehicles, real-time and precise estimation of object distances is crucial for safe and efficient navigation \cite{zhang2020regional,agand2023deep,yang2019reactive,kang2019df}. Traditionally, distance estimation is performed using stereo or multi-camera imaging systems or LiDAR measurements, both of which have their own limitations that can impact their use cases and scalability. Stereo imaging requires precise synchronization between two cameras, which can introduce multiple points of failure. Furthermore, stereo vision is limited by the distance between the cameras and the texture of the region of interest \cite{saxena2007depth}. Although accurate, LiDAR systems are considerably more expensive to purchase and operate than a single camera. Moreover, they have several moving parts and components that can fail, and equipping a vehicle with multiple LiDAR devices for redundancy is prohibitively expensive \cite{lidar_safety}. In contrast, a system that uses a single camera can incorporate several backup cameras for the price of a single LiDAR device, making it more cost-effective and scalable.

However, existing monocular object distance estimation techniques suffer from accuracy issues or labor-intensive data collection requirements. Monocular vision has inherent difficulties in accurately estimating object distances, and current solutions typically involve a combination of a 2D object detector and a monocular image depth estimator or a monocular 3D object detector \cite{svr}. The former approach relies heavily on a monocular depth estimator that is not optimized for precise object-wise depth estimation, while the latter requires additional annotations of 3D bounding box (BBox) coordinates for training, resulting in specialized equipment and high labeling costs. Consequently, there is a need for a reliable and cost-effective approach that can be easily adapted to new settings and object detectors.

Numerous studies have investigated the use of deep neural networks (DNN) for direct object distance estimation. Early approaches such as inverse perspective mapping (IPM) \cite{TuohyIPM} converted image points  into bird's-eye view coordinates. However, IPM has limitations, especially for distant objects (around 40m) or non-straight roads \cite{zhu2019learning}. Other unsupervised methods include learning from unstructured video sequences \cite{zhou2017unsupervised}, employing CNN feature extractors with distance and keypoint regressors \cite{zhu2019learning}, and modifying MaskRCNN as demonstrated in \cite{zhang2020regional}. Additionally, a  self-supervised framework for fisheye cameras in autonomous driving was  enhanced with a multi-task learning strategy \cite{kumar2021syndistnet}. Authors in \cite{shi2023structured} proposed an end-to-end approach called structured convolutional neural feld (SCNF) that combine CNN and continuous condition random field.

The accuracy of class-specific object detection relies on matching the training environment \cite{agand2023online}. For example, in a test scenario involving toy objects, a toy car at the same distance as a real car will appear much smaller but may still be detected as a ``Car'' by object classification networks. Similarly, when an object is presented in the camera field of view while tilted, a class-specific approach can only detect the distance correctly if the object is in the dataset with the exact pose, which is unlikely or requires an enormous dataset. Finally, the precision in class-specific approaches with multiple classes is limited to the accuracy of the classification technique. These limitations don't affect class-agnostic approaches \cite{agand2019adaptive}, which don't require knowledge of expected object sizes at varying distances.

\pa{In this paper, we introduce DMODE, a novel approach to object distance estimation that addresses significant challenges. By avoiding reliance on object class information, we prevent the model from memorizing object size patterns at various distances. Instead, DMODE utilizes changes in an object's projected size over time and camera motion for distance estimation. Our approach achieves three primary contributions: 1) It provides accurate distance estimations without requiring object class information, overcoming the challenge of class-agnostic estimation. 2) It is independent of camera intrinsic parameters, ensuring adaptability to diverse camera setups. 3) It is able to generalize and accurately estimate distances for unseen object classes, enables efficient transfer learning for new ones, and addresses the challenge of adaptability across different object tracking networks (OTN) and deployment scenarios. To facilitate future studies, the code is available in GitHub at \url{https://github.com/pagand/distance-estimation} 
}


\section{Related work}

\subsection{Monocular depth estimation}
Depth estimation has been approached using DNNs, such as continuous condition random fields for image patches proposed by \cite{liu2015learning}. The accuracy of depth estimation was improved by \cite{fu2018deep}, who incorporated ordinal regression into the depth estimation network and used scale-increasing discretization to convert continuous depth data into an ordinal class vector. These techniques require significant manpower and computing resources, as well as specific training images and corresponding depth maps for each pixel \cite{zhang2020monocular}. In the absence of a depth image serving as ground truth (GT), unsupervised training might utilize additional depth cues from stereo images \cite{godard2017unsupervised} or optical flows \cite{godard2019digging}. Authors in \cite{wan2022multi} introduced a novel deep visual-inertial odometry and depth estimation framework to enhance the precision of depth estimation and ego-motion using image sequences and inertial measurement unit (IMU) raw data. However, unsupervised depth estimation methods have inherent scale ambiguity and poor performance due to the lack of perfect GT and geometric constraints \cite{wan2022multi}.


\subsection{Monocular 3D object detection}
The challenging task of 3D object recognition from monocular images is related to object distance estimation. Mousavian \etal~\cite{mousavian20173d} proposed Deep3DBox, which employs a 3D regressor module to estimate the 3D box's dimensions and orientation and takes 2D detection input to crop the input features. To replace the widely used 2D R-CNN architecture, \cite{brazil2019m3d} introduced a 3D region proposal network, significantly improving performance. Furthermore, some studies use point cloud detection networks or monocular depth estimation networks as supplementary components of monocular 3D object recognition \cite{you2019pseudo}. These additional details enhance the accuracy of 3D object detection networks.


\subsection{Monocular object distance estimation}
Ali and Hussein \cite{ali2016distance} used a geometric model incorporating camera characteristics and vehicle height as inputs to determine the distance between two cars. Bertoni \etal~\cite{bertoni2019monoloco} employed a lightweight network to predict human positions from 2D human postures.  A generic object distance estimation was developed by adding a depth regression module to the Faster R-CNN-based structure and an additional keypoint regressor to improve performance for objects near the camera \cite{zhu2019learning}. Cai \etal~\cite{cai2020monocular} proposed a framework that breaks down the problem of monocular 3D object recognition into smaller tasks, including object depth estimation, and introduced a height-guided depth estimation technique to address the information loss caused by the 3D to 2D projection. An R-CNN-based network was used to achieve object recognition and distance estimation simultaneously \cite{zhang2021regional}.

\begin{figure}[t]
\centering
    \includegraphics[width=0.7\linewidth,trim={3cm  5cm 7cm 3cm},clip]{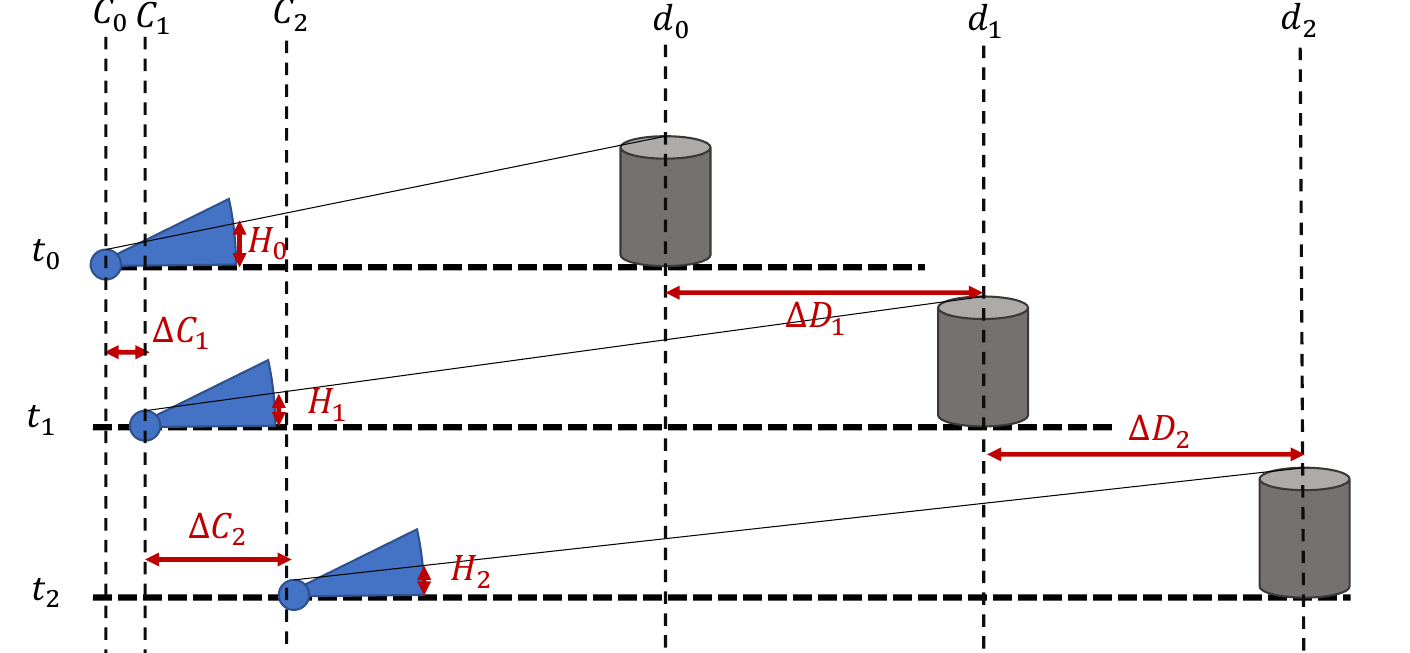}
\caption{Simplified 1D DMODE: a mathematical viewpoint}
\label{fig:alg}
\end{figure}

\section{Problem statement}
Our goal is to estimate object distances in 3D by determining their relative Cartesian positions to a single camera without using any class-related information. The camera can be mounted on a moving vehicle or be stationary. 
\pa{For the sake of simplicity,} Fig. \ref{fig:alg} depicts \pa{an illustrative scenario involving camera motion in 1D} for three time instants, $t_0, t_1, t_2$. Let $d_j$ be the object's distance from the camera, $D_j$ be the object's absolute position, $H_j$ the object's size as captured by the camera, and $C_j$ the camera's absolute position at time $t_j$.  We define $\Delta F_j =  F_j - F_{j-1}$, where $F$ can be any of the aforementioned variables. 
Our objective is to compute $d_2$, the object's relative position with respect to the camera at the current (latest) time, given  $H_0, H_1, H_2, \Delta C_1, \Delta C_2$. 

Here are the assumptions: 1) The distance between the object and the camera in the captured frames varies, which can be caused by the movement of the object, camera, or both. 2) The camera's movement is measured by an IMU \cite{agand2017teleoperation}. 3) \pa{Within the captured frames, the object does not have a pitch rate (rotation around a horizontal axis perpendicular to the line connecting the camera and the object)}.

\section{Method}
\begin{figure*}[t!]
    \centering
    \includegraphics[width=0.8\linewidth,trim={.5cm  12.5cm 0cm 0cm},clip]{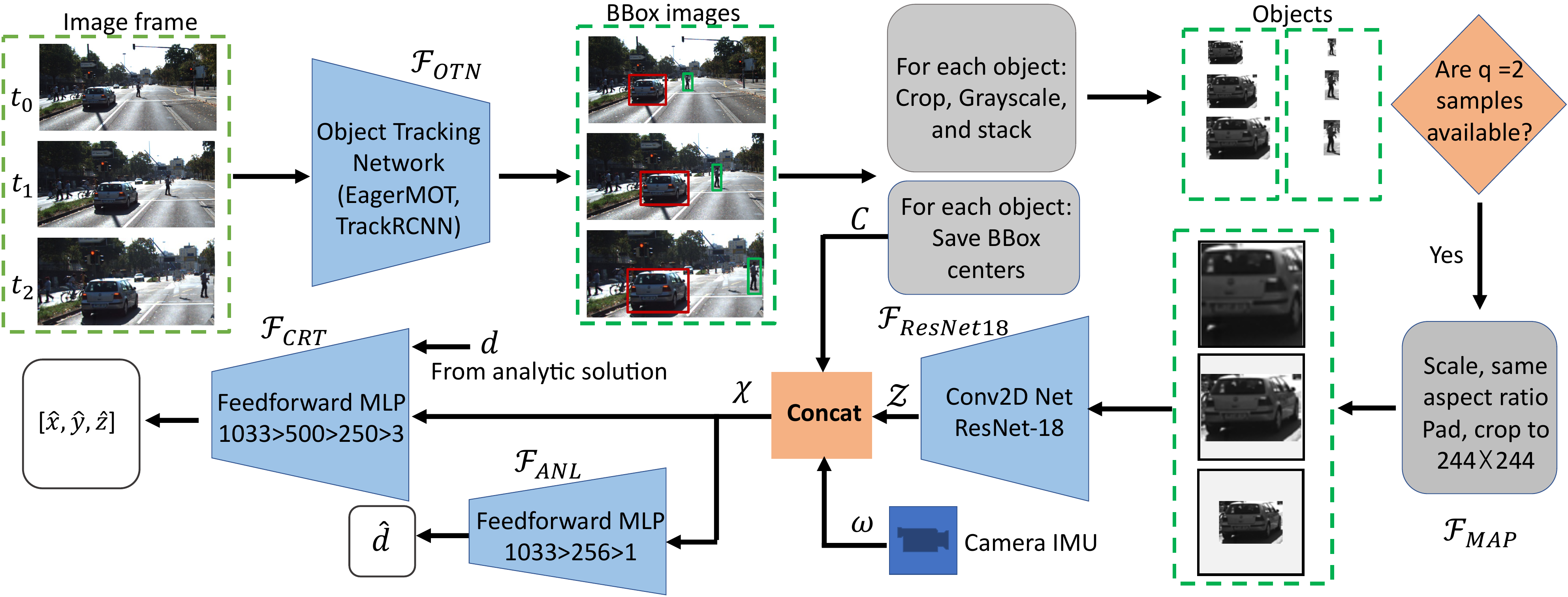}
    \caption{DMODE framework: workflow of data and models. Only the blue color elements are trainable. The green dots show stack of data. }
    \label{fig:Workflow}
\end{figure*}

The framework is depicted in Figure \ref{fig:Workflow}. Our method involves tracking the projected size of an object in the camera lens over a predefined time frame while taking into account the camera's motion to estimate the object's distance from the camera. 
To achieve this, we require an OTN and an IMU for the camera. The distance to the object with constant velocity can be analytically computed using the following relation:
\begin{equation}
d = \frac{H_0H_1(\Delta C_1 - \Delta C_2)}{H_2 \Delta H_1 - H_0\Delta H_2},
    \label{analytic}
\end{equation}
where $H_i, C_i$ are the object pixel height and camera location at $i$-th sequence, and $\Delta X_i = X_i - X_{i-1}$ for $X \in \{H, C\}$. \pa{In Sec. IV-A, to demonstrate the model-agnosticity,  we derive a mathematical expression for calculating the distance to a 3D object with any dynamic movement of order $q$ and unknown parameters. This analytical proof will illustrate that, to determine the distance to an object with sufficient frames (i.e., $q+1$ frames), there is no requirement for information about the object's class or shape. In reality, all the required variables in Eq. (\ref{analytic}) are measured with noise. The inconsistency between the height measurements is limited by the accuracy of the bounding box and the OTN. Therefore in Sec. IV-B and C, we propose an architectural design and learning framework that makes the use of bounding boxes and ego motion for depth estimation more robust.}  

\subsection{Theoretical analysis}
By extending Fig. \ref{fig:alg} to 3D, we aim to determine a function that outputs the relative coordinates of the object $\phi = (x, y, z)$ with respect to the camera. 
For convenience, suppose $\Delta t = t_2 - t_1 = t_1 - t_0$ and $C_0=0$. 
\begin{equation}
    \phi = \mathcal{F}(H_1, H_2, \ldots, H_{q+1}, \Delta C_1, \Delta C_2, \ldots, \Delta C_q),
\end{equation}

\noindent where $q+1$ is the number of input frames and $\mathcal{F}$ is an unknown function need to be determined. 
To alleviate the effects of noisy measurements and potential errors in OTN, we utilize a DNN to derive $\mathcal{F}$. The distance to the object is calculated using $d = \sqrt{x^2+ y^2+ z^2}$.
For the sake of illustration, we derive the distance for the simplified 1D scenario as shown in Fig. \ref{fig:alg}.\\
Axiom: Object projected image size and the distance to camera are inversely proportional \cite{alphonse2021depth}
\begin{equation}
	\frac{d_{n-1}}{d_n} =\frac{H_n}{H_{n-1}} \doteq	P_n
	\label{eq:ax1}
	\end{equation}
	
\noindent Theorem: Given $q+1$ frames, the following relations apply for all $i \in \{1, 2, \cdots,q\}$:
\begin{equation}
    (1-p_i) \sum_{k=i+1}^{q+1} \Delta D_k + \Delta D_i = \sum_{k=i}^q \Delta C_k - p_i \sum_{k=i+1}^{q} \Delta C_k
\label{eq:theorem}
\end{equation}
where $d_j = \Delta C_j = 0, \forall j>q$.

\begin{proof}
According to Fig. \ref{fig:alg}, we have:
\begin{equation}
d_{n}=d_{n-1}+\Delta D_{n}-\Delta C_{n}
\label{eq:111}
\end{equation}
First let us prove the last relation ($i=q$) in (\ref{eq:theorem}). By using Eq. (\ref{eq:ax1}) for $n=q$, we have $d_{q-1} = p_q d_q$, which  can be substituted in (\ref{eq:111}) for $n=q$.
This proves the  last relation in (\ref{eq:theorem}) ($i=q$), since $\Delta D_{q+1}=d_q$, we have:
\begin{equation}
p_q = \frac{d_q-\Delta D_q +\Delta C_q}{d_q}
    \label{eq:pq}
\end{equation}
Now for $i=q-1$, we can write similar relation to (\ref{eq:pq}) as follows by considering the previous time step and substituting $d_{q-1}, d_{q-2}$ using  (\ref{eq:ax1}) for $n=q, q-1$:
\begin{equation}
d_qp_q=p_{q-1}d_qp_q+\Delta D_{q-1}-\Delta C_{q-1}
\label{eq:333}
\end{equation}
Using the relation for $p_q$ in (\ref{eq:pq}), we can simplify the relation in (\ref{eq:333})  as follows:
\begin{equation}
    (d_q - \Delta D_q + \Delta C_q)(1-p_{q-1}) = \Delta D_{q-1} - \Delta C_{q-1}
\end{equation}
which is equivalent to the second last relation in (\ref{eq:theorem}) ($i=q-1$). The proof for other relations follows the same pattern. Now we need to integrate the assumption about object movement into $\Delta D_j$ variables to reduce one of them as unknown variable (i.e for object with constant velocity we have $\Delta D = V \Delta t$ or with constant acceleration, $\Delta D = 1/2 a \Delta t^2+ V \Delta t$). The general relation is as follows:
\begin{equation}
\underbar{A}(P)[d_q, \underbar{f}]^T = \underbar{b}(C, P)
    \label{eq:gen}
\end{equation}
where $\underbar{A}$ is a $q\times q $ matrix of $(P_1, \cdots, P_q)$ parameters, $\underbar{f}$ is a row vector of relative movement components (e.g. velocity, acceleration, etc), and $\underbar{b}$ is a column vector of $(P_1, C_1, \cdots, P_q, C_q)$ parameters. The analytic solution to derive distance in Eq. (\ref{eq:gen}) is given by:
\begin{equation}
d_q =\underbar{A}^{-1}_1 \underbar{b}(C, P) 
\end{equation}
where $ A^{-1}_1$ is the first row of $A$ inverse.
\end{proof}
For a special case, where  the objects of interest is stationary ($q=1$), the relation will be simplified to:
\begin{equation}
(1-p_1)d_1 = \Delta C_1, ~~ \textit{if}~ q =1
\end{equation}
By assuming objects moving with constant velocity ($q=2$), we have $\Delta D_1 = \Delta D_2 = \Delta D$ and  Eq. (\ref{eq:gen})  is updated as:
\begin{equation}
\begin{bmatrix} 
P_1-1&2-P_1 \\
	P_2-1 & 1 \\
	\end{bmatrix} \begin{bmatrix}d_2\\
	\Delta D
	\end{bmatrix} =\begin{bmatrix} 
\Delta C_1+\Delta C_2-P_1 \Delta C_2 \\
\Delta C_2 \\
	\end{bmatrix}
\end{equation}
Where we can solve for $d_2$ and $\Delta D$. 
\begin{equation}
d_2 =\frac{\Delta C_1-\Delta C_2}{P_1P_2-2P_2+1}    ,~~ \textit{if}~ q = 2
\label{eq:anlytic}
\end{equation}
This means that analytically, if the object is moving with constant velocity, the camera should not be moved with a constant velocity. 
However, as mentioned before, upon inspection of the GT annotations, the change of an object's BBox size is often inconsistent with its distance changes from the camera.  Replacing $P_1$ and $P_2$ with it's definition given in Eq. (\ref{eq:pq}), proves Eq. (\ref{analytic}).

\subsection{Network architecture}


\pa{As shown in Fig. \ref{fig:Workflow}}, the sequential network, inspired from \cite{fan2024sequential}, consist of a 2D convolutional neural network (CNN) and three multi-layer perceptrons (MLPs) with batch normalization implemented between each layer. After processing the detected object from OTN ($\mathcal{F}_{OTN}$), three consecutive frames are stacked and passed through a mapping function ($\mathcal{F}_{MAP}$) comprised of an MLP with a soft-plus activation function. \pa{As shown in the upper right corner of Fig. \ref{fig:Workflow}, upon the availability of 3 frames of an object, we resize and crop them to the size of $224\times 224$}. We employ ResNet-18 ($\mathcal{F}_{ResNet18}$) \cite{he2016deep} by inputting selected  bounding box (BBox) information to yield $1000$ parameters as latent variables ($\mathcal{Z}$).

\begin{equation}
    \begin{split}
       &B_{t_i} = \mathcal{F}_{OTN} (I_{t_i}); ~~ \forall i \in \{0,1,2\}, \\
       &S_{t} = \mathcal{F}_{map} (B_{t_1}, B_{t_2}, B_{t_3}   ), \\
      &\mathcal{Z}_{(1\times 1000)} =\mathcal{F}_{ResNet18}(S_{t}), \\
    \end{split}
    \label{eq:arc}
\end{equation}

\noindent where, $I_{t_i}$ and $B_{t_i}$ represent the image frame and the BBox information in the time frame $t_i$, respectively. Additionally, $S_{t}$ is the stacked processed BBox image after applying the mapping. The latent variables are then concatenated with an additional 33 variables to create BBox features. These variables encompass the camera's Cartesian velocities, accelerations, and angular accelerations of the three keyframes ($\omega_{(1\times 27)}$). The remaining parameters are the object's BBox centers relative to the image dimensions ($C_{(1\times 6)}$), which are the only BBox information not related to their class. The BBox features ($\mathcal{X}_{(1\times 1033)} = [\mathcal{Z},C, \omega]$) are fed to another MLP to derive three separate parameters in Cartesian space ($[x,y,z] =\mathcal{F}_{CRT}(\mathcal{X},d)$). To enforce the computed analytical solution for $d$, we use a separate head with an MLP and ReLU to enrich the BBox features ($\hat{d} =\mathcal{F}_{ANL}(\mathcal{X})$). \pa{ The reason for predicting Cartesian space separately is to allow the model to distinguish between different spatial coordinates and ultimately enhance the overall performance. Additionally, directly learning $d$ with $\mathcal{F}_{ANL}$ will update the bottleneck node ($\mathcal{X}$) in the backpropagation direction. }






\subsection{Learning rules}


Upon object detection in any frame, the cached data is examined. If detection data for this object exists for all keyframes, it is prepared by $\mathcal{F}_{map}$ for the following stages: The system crops the detected objects along their BBoxes and converts them into single-channel grayscale images. For each object, the system collects the corresponding images, and BBox information for the keyframes. The triplet of images is then resized relative to the heights of their BBoxes while maintaining their original aspect ratios. The largest image among the three is scaled to a height of 224, and all three images are padded/cropped to a size of $224\times224$. Subsequently, they are overlaid at the center in the order of detections, forming a new 3-channel image that serves as the 2D input to $\mathcal{F}_{ResNet18}$. It is important to note that $\mathcal{F}_{OTN}$ in Eq. (\ref{eq:arc}) is a pre-trained network, $\mathcal{F}_{map}$ is an arithmetic manipulation, and only $\mathcal{F}_{ResNet18}$, $\mathcal{F}_{CRT}$, and $\mathcal{F}_{ANL}$ are trained concurrently. The objective function is based on the robust Adaptive Reverse Huber Loss (BerHu) \cite{lambert2016adaptive}:

\begin{equation}
    B(\phi,\phi^*)= \begin{cases}
        |\phi-\phi^*|, & \text{if } |\phi-\phi^*| \leq c,\\
        \frac{(\phi-\phi^*)^2+c^2}{2c}, & \text{otherwise } 
        \end{cases} 
\end{equation}
\noindent where, $\phi^*$ represents the corresponding GT of Cartesian space, and $c = 0.2 . \max{(|\phi-\phi^* |)}$. The BerHu loss function integrates $L_1$ and $L_2$ features and adapts according to the $c$ value. Contrary to the $L_2$ loss function, the BerHu loss function converges more rapidly while effectively preserving small residuals, as it is differentiable at the point $c$ where the transition between $L_1$ and $L_2$ losses takes place. This enables the network to leverage the shrinkage power of $L_1$ norm, thereby avoiding low gradients for relatively small residuals. 
The final loss inspired from multi-task training in \cite{agand2023letfuser} is defined as follows:

\begin{equation}
L = \frac{1}{4N}\sum_{i=1}^N ( B(x_i,x_i^*)+B(y_i,y_i^*)+B(z_i,z_i^*) + B(d_i, d_i^*) ),
\end{equation}
\noindent where $N$ represents the number of mini-batch samples, and $i$ is the index of all images for the current batch. We utilized the ADAM optimizer with a weight decay of 1e-5.   To prevent over-fitting, we implemented dropout and batch normalization approach \cite{agand2023fuel,garbin2020dropout}.

\section{Results}

\subsection{Model setup}

Our model capitalizes on the evolution of an object's size over time, paired with camera movements, to estimate distances. We employ a 3-interval scheme, mathematically presuming constant velocity. To discern a noticeable difference, we have chosen a 1-second look-back time frame. The keyframes for this scheme will be frames $n$, $n-5$, and $n-10$, corresponding to $t_2$, $t_1$, and $t_0$, respectively, given that KITTI's camera records at 10 frames per second. For our dataset, we utilize the images, annotations, and IMU data from the KITTI Multi Object Tracking dataset \cite{milan2016mot16}. This dataset includes detection, classification, and tracking data for seven categories.
We omit object detections labeled as ``Misc'' or ``DontCare'' from the KITTI dataset. Furthermore, we filter out all entries where the object is not entirely within the frame, and disregard object detections with bounding boxes touching the frame's edge during testing. To ensure consistency across images, occluded object detections are not included in the training process. The size of objects is determined using the height of their bounding boxes (BBoxes). 
Our training and validation split adheres to the data separation employed by TrackRCNN (TRCNN) and EagerMOT (EMOT). We base our results on the validation set, with each model trained over 100 epochs, a batch size of 32, and a learning rate of 1e-3. We estimate distances to the centers of objects, as defined by KITTI's 3D BBox annotations, rather than the point closest to the camera. 

To evaluate our method, we use the following common metrics. Each metric has a corresponding mark where  $\downarrow$ means ``lower is better'' and $\uparrow$ means ``higher is better''.  1) Median relative error $MRE \downarrow = \text{Median}([\left \|d_1^*-d_1\right \|, ..., \left \|d_n^*-d_n\right \|])$ 2) Absolute relative error $\text{AbsRel} \downarrow =  \frac{1}{N}\sum \left ( \frac{ \left \|d_i^*-d_i\right \| }{d_i} \right )$ 3) Square relative error $\text{SquirRel} \downarrow = \frac{1}{N}\sum \left ( \frac{ \left \|d_i^*-d_i\right \|^2 }{d_i} \right )$ 4)  Threshold $\delta \uparrow = \max(d_i/d_i^*, d_i^*/d_i)\% < 1.25$ 5) confidence interval $ .95\text{CI} \downarrow  = Rel. Err. \pm x $ 6) Root mean square error $\text{RMSE} \downarrow = \sqrt{\frac{1}{N} \sum \left \|d_i^*-d_i\right \|^2}$   7) $\text{RMSE}\textsubscript{log} \downarrow = \sqrt{\frac{1}{N} \sum \left \|\log d_i^* - \log d_i\right \|^2}$

To assess our method's robustness across diverse data sources, we tested it using  TRCNN \cite{voigtlaender2019mots} and EMOT \cite{kim2021eagermot}. However, a limitation arises from these OTNs, which only detect cars and pedestrians, while KITTI annotations encompass seven object classes. The unique identifiers assigned by the OTN sometimes differ from those in the GT annotations. To assign GT distances to OTN-detected objects in each frame, we compare the OTN's BBox outputs with GT BBoxes. The GT annotation is assigned to one of the OTN's outputs if it maximizes the following formula among all detections ($BOX_{pred}\cap BOX_{GT}-BOX_{pred}\oplus BOX_{GT}$).

\begin{table*}
\centering
\caption{Comparison between baselines and our method.}
\begin{tabular}{ccccccc}
  \toprule
Class type&Method& $\delta < 1.25 \uparrow$ &AbsRel $\downarrow$ & SquirRel $\downarrow$ & RMSE $\downarrow$ & RMSE\textsubscript{log} $\downarrow$\\
\midrule
\multirow{6}{*}{Car only}&SVR \cite{svr} & 0.345 & 1.494 & 47.748 & 18.970 & 1.494\\
&IPM  \cite{TuohyIPM}& 0.701 & 0.497 & 1290.509 & 237.618 & 0.451\\
&Enhanced res50 \cite{zhu2019learning}& 0.796 &  0.188 & 0.843 & 4.134 & 0.256\\
&Enhanced vgg16 \cite{zhu2019learning} & 0.848 & 0.161 & \textbf{0.619} & \textbf{3.580} & 0.228\\
&Ours  & \textbf{0.849} & \textbf{0.107} & 0.998 & 4.931 & \textbf{0.173}\\
\midrule
\multirow{6}{*}{Pedestrian only}&SVR  \cite{svr}& 0.129 & 1.499 & 34.561 & 21.677 & 1.260\\
&IPM \cite{TuohyIPM} & 0.688 & 0.340 & 543.223 & 192.177 & 0.348\\
&Enhanced res50  \cite{zhu2019learning} & 0.734  & 0.188 & 0.807 & 3.806 & 0.225\\
&Enhanced vgg16  \cite{zhu2019learning} & 0.747 & 0.183 & 0.654 & 3.439 & 0.221\\
&Ours  & \textbf{0.842} & \textbf{0.109} & \textbf{0.635} & \textbf{3.110} & \textbf{0.149}\\
\midrule
\multirow{6}{*}{Average on Muliple classes}
& Zhou (ResNet-50) \cite{zhou2017unsupervised} &0.678 &  0.208& 1.768& 6.856& 0.283 \\
&Enhanced  res50 \cite{zhu2019learning} & 0.550 & 0.271 &  2.363 & 8.166 & 0.336\\
&Enhanced vgg16  \cite{zhu2019learning} & 0.629 & 0.251 & 1.844 & 6.870 & 0.314\\
& SynDistNet (ResNet-50) \cite{kumar2021syndistnet}&\textbf{0.896} &  0.109& \textbf{0.718}& 4.516&0.180 \\
& SCNF \cite{shi2023structured}&0.775 &  0.184& 0.842& 9.773&0.311 \\
&Ours & 0.847 & \textbf{0.108} & 0.872 & \textbf{4.382} & \textbf{0.165}\\
\bottomrule
\end{tabular}
\label{Tab:comp}
\end{table*}

\subsection{Comparison}

We compared our method with baselines in Table \ref{Tab:comp}. Our model was trained using EMOT as the OTN on the full set of GT annotations for all evaluation scenarios. Among recent literature, only SVR \cite{svr}, IPM \cite{TuohyIPM}, and Enhanced res50/vgg16 \cite{zhu2019learning} provided their single-class evaluations. The accuracy measurements for IPM and SVR were taken from Zhu and Fang experiments \cite{zhu2019learning}. In single-class evaluations for cars, our model performed better in AbsRel, threshold, and RMSE\textsubscript{log}, while performing worse in SquirRel and RMSE. This suggests that while our model is more accurate on average, it exhibits more extreme outliers. For pedestrians, our approach surpassed all other methods that provided their single-class evaluation. In terms of multi-class training, our model outperformed all of these methods across all metrics. We not only surpassed the performance of \cite{zhu2019learning}'s non-class-specific model but also outperformed their model that utilized a classifier. We additionally include two other approaches: SCNF \cite{shi2023structured} and SynDistNet (ResNet-50) \cite{kumar2021syndistnet}, both of which only provided their multi-class evaluation. SynDistNet results were obtained from an experiment conducted by Liang \etal~\cite{liang2022self}. Although our approach is designed to perform well with under-specified or unknown classes, it still outperforms these approaches in terms of AbsRel, $\text{RMSE}$ and $\text{RMSE}_{\log}$ for the KITTI dataset and performs comparably in other metrics. We interpret this performance as resulting from high penalties applied to objects too far away, where their relative changes in two frames are not noticeable.

\section{Ablation studies}
\subsection{Dataset robustness testing}
Table \ref{tab:class} presents the model's performance when tested on the full annotation of the validation set. To evaluate our method's robustness regarding the dataset, we assessed the model's performance when fed different sources of detections of the validation set. The model was initially trained on the full annotations of the GT training set. As EMOT and TRCNN only detect ``Cars'' and ``Pedestrians'', we included the class-limited GT results as a baseline ($GT^*$). Furthermore, to evaluate the model's ability to transfer learning across different object detection systems, we present our model's performance when it was trained and tested on TRCNN's outputs, as shown in the last row of Table \ref{tab:class}.

\begin{table}[ht]
\caption{Model trained and tested on GT or OTN}
\begin{center}
\noindent
\begin{tabular}{cccccc}
\toprule
  Train type&Test type & AbsRel $\downarrow$ & MRE $\downarrow$ & .95CI $\downarrow$ & RMSE $\downarrow$\\
\midrule
 GT&GT & 0.145 & \textbf{0.107} & 3.9e-3 & 6.765\\
GT&$GT^*$ & \textbf{0.138} & \textbf{0.107} & \textbf{3.2e-3 }& \textbf{6.676}\\
 GT &TRCNN & 0.168 & 0.125 & 4.6e-3 & 7.139\\
GT &E-MOT & 0.160 & 0.124 & 3.7e-3 & 6.435\\
TRCNN&TRCNN & 0.163 & 0.118 & 4.8e-3 & 7.202\\
\bottomrule
\end{tabular}
\end{center}
\label{tab:class}
\end{table}

\begin{table}[ht]
\caption{Models tested on GT with car only class }
\begin{center}
\noindent
\begin{tabular}{cccccc}
\toprule
 Model &Train class &  AbsRel $\downarrow$ & MRE $\downarrow$ & .95CI $\downarrow$ & RMSE $\downarrow$\\
\midrule
 Enhanced  &All & 0.148 & \bf{0.101} & \bf{3.8e-3} & \bf{5.953}\\
vgg16&No Car & 0.578 & 0.534 & 9.8e-3 & 21.773\\
\cite{zhu2019learning} &TL & 0.307 & 0.213 & 5.6e-3 & 10.487\\
\midrule
 \multirow{3}{*}{Ours}  &All & \bf{0.143} & 0.113 & 4.1e-3 & 7.821\\
&No Car & \bf{0.282} & \bf{0.243} & \bf{6.6e-3} & \bf{13.923}\\
 &TL & \bf{0.173} & \bf{0.133} & \bf{4.9e-3} & \bf{9.356}\\
 \bottomrule
\end{tabular}
\end{center}
\label{tab:car}
\end{table}

\subsection{Class agnostic testing}
\pa{As the employed OTN only tracks cars and pedestrians, and to make our results comparable to other literature, we attempted to test class agnosticism by removing the ``Car'' class from the dataset and evaluating its performance during testing. Table \ref{tab:car} demonstrates our method's performance in estimating the distances of cars, even though they were not part of the training set. It also illustrates the accuracy improvement through transfer learning (TL) when the previously missing class is included in the training data. Therefore, our approach provides more accurate results for classes that were never encountered in the training dataset compared to non-model-agnostic rivals that provide better individual class predictions for known classes. }

\begin{figure}[t!]
    \centering
    \includegraphics[width=0.8\linewidth,trim={2cm  1.2cm 1cm 1.5cm},clip]{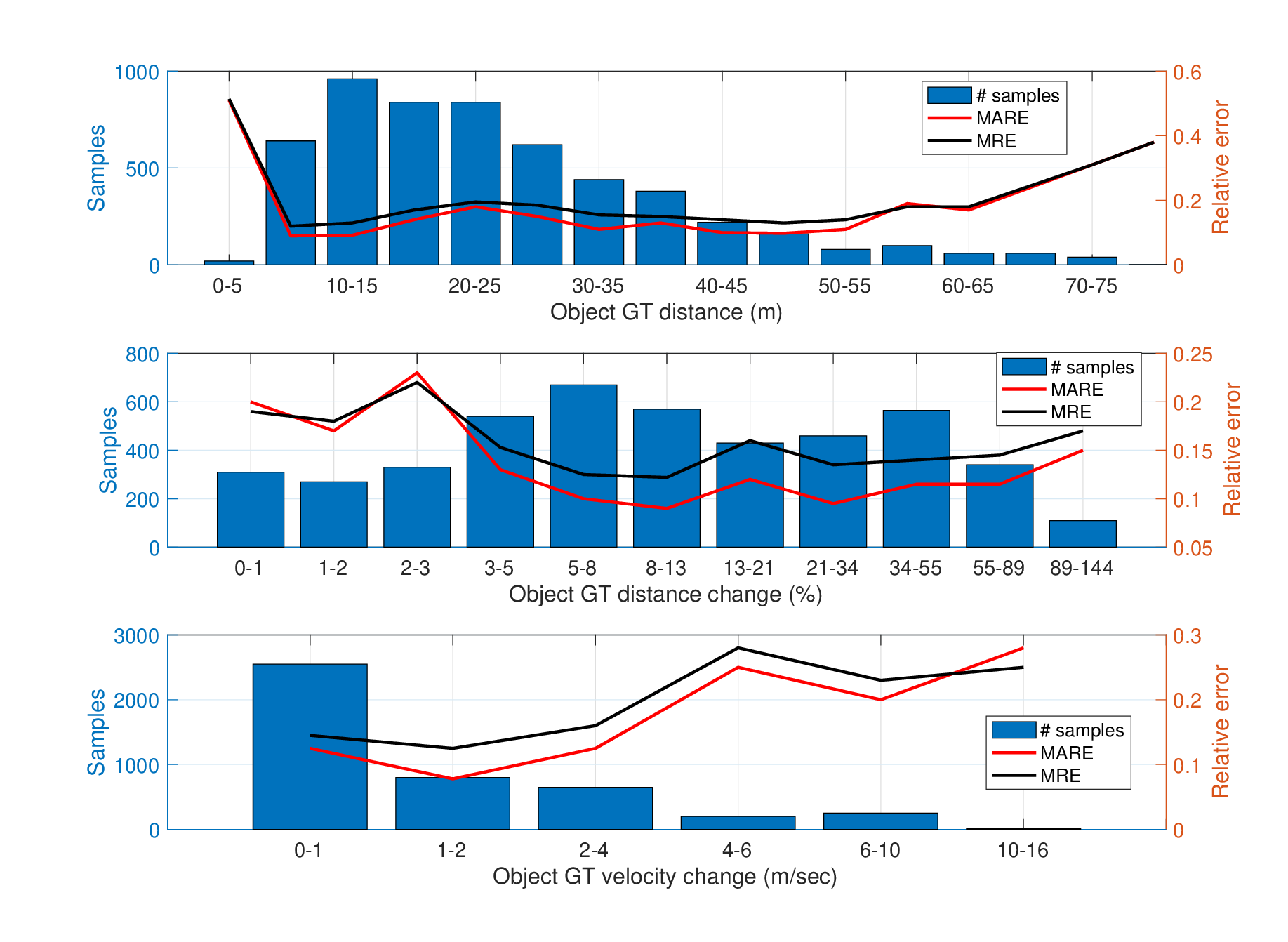}
    \caption{Error bar in relation to distances, velocity, and acceleration. The red and black lines respectively represents AbsRel and  MRE.}
    \label{fig:bar}
  \end{figure}
  
 \subsection{Discussion}
\pa{Often, a trade-off exists between versatility and specialization. Class-agnostic methods excel in versatility, accommodating a wide array of objects without requiring knowledge of specific classes. This adaptability is crucial in scenarios with diverse and unexpected objects. Conversely, specialized class-aware methods may perform admirably in specific situations but could struggle with unfamiliar objects or scenarios.}

Dataset robustness show that our method, trained on GT annotations, can perform well with various OTNs, with minimal accuracy loss. Additionally, training the model on the specific OTN outputs used during testing yields similar accuracy to training on GT annotations. These results indicate our model's suitability for deployment across diverse systems with different OTNs, maintaining high accuracy without the need for retraining. In class-agnostic testing, our method showed a notably smaller accuracy decrease (97\% increase in AbsRel) for unseen object classes compared to Enhanced vgg16 (290\% increase in AbsRel), indicating our model's reduced dependency on class-specific training data.


Fig. \ref{fig:bar} displays error bars in a histogram for different object distances, velocities, and accelerations. Our method excelled for objects within the 5 to 60-meter range, where most detections occurred. The error spike within 5 meters is due to limited training samples, as truncated objects were excluded from training. Beyond 60 meters, error increased due to camera resolution limitations and challenges in representing size changes in the 2D input. The analysis of the second subplot in Fig. \ref{fig:bar} indicates our model's effectiveness increases for objects with significant relative distance changes from the camera compared to those with minimal changes, resulting in a noticeable 3-5\% accuracy improvement. However, it's important to note that the 3-interval version of our method is only suitable for objects moving at a constant velocity, as discussed in the mathematical motivation section. In the third subplot, we find that our model maintains accuracy for objects with velocity changes below 2m/s but experiences a decline in performance for objects with greater motion variations. Fortunately, most objects in scenarios like city driving, as captured in the KITTI dataset, do not exhibit velocity changes exceeding 2m/s. Consequently, our current model provides distance estimations with an average relative error of 13.5\%, despite its limitations.



\section{Conclusion}

We introduced a class-agnostic method for object distance estimation, leveraging the object's evolving appearance and camera motion. This property enables TL with target class samples, yielding satisfactory multi-class distance estimation. The limitation of our approach includes  time delay and accurate object tracking over three key frames, matching the training intervals, is essential. Moreover, The network architecture currently employs three input channels, suitable for constant object velocity. While results are generally good, outliers may occur for rapidly changing objects. Future work may involve increasing the channel capacity of our 2D network to improve performance in such cases. 

\bibliographystyle{IEEEtran}
\bibliography{root}

\end{document}